\documentclass[letterpaper, 10 pt, journal, twoside]{IEEEtran}
\usepackage{amsmath,amsfonts}
\usepackage{algorithmic}
\usepackage{algorithm}
\usepackage{array}
\usepackage[caption=false,font=normalsize,labelfont=sf,textfont=sf]{subfig}
\usepackage{textcomp}
\usepackage{stfloats}
\usepackage{url}
\usepackage{verbatim}
\usepackage{graphicx}
\usepackage{cite}
\usepackage{multirow} 
\usepackage{booktabs}
\usepackage{xcolor}

\title{\LARGE \bf
Why Not Replace? Sustaining Long-Term Visual Localization via Handcrafted-Learned Feature Collaboration on CPU
}

\author{Yicheng Lin, Yunlong Jiang, Xujia Jiao and Bin Han, \textit{Senior Member, IEEE}
\thanks{This work was supported in part by the National Natural Science Foundation of China (52375015) and in part by the Natural Science Foundation of Hubei Province of China (2022CFB239). (Corresponding author: Bin Han)}
\thanks{Y. Lin, Y. Jiang, X. Jiao and B. Han are with the State Key Laboratory of Intelligent Manufacturing Equipment and Technology, School of Mechanical Science and Engineering, Huazhong University of Science and Technology, Wuhan 430074, China (e-mail: {\{\tt\small yichenglin, jiangyunlong, xujiajiao and binhan \}@hust.edu.cn}).}
\thanks{Digital Object Identifier (DOI): see top of this page.}%
}

\begin{document}
\maketitle

\begin{abstract}
Robust long-term visual localization in complex industrial environments is critical for mobile robotic systems. Existing approaches face limitations: handcrafted features are illumination-sensitive, learned features are computationally intensive, and semantic- or marker-based methods are environmentally constrained. Handcrafted and learned features share similar representations but differ functionally. Handcrafted features are optimized for continuous tracking, while learned features excel in wide-baseline matching. Their complementarity calls for integration rather than replacement. Building on this, we propose a hierarchical localization framework. It leverages real-time handcrafted feature extraction for relative pose estimation. In parallel, it employs selective learned keypoint detection on optimized keyframes for absolute positioning. This design enables CPU-efficient, long-term visual localization. Experiments systematically progress through three validation phases: Initially establishing feature complementarity through comparative analysis, followed by computational latency profiling across algorithm stages on CPU platforms. Final evaluation under photometric variations (including seasonal transitions and diurnal cycles) demonstrates 47\% average error reduction with significantly improved localization consistency. The code implementation is publicly available at https://github.com/linyicheng1/ORB\_SLAM3\_localization.
\end{abstract}

    
\section{Introduction}

\IEEEPARstart{O}{ver} the past few decades, the development of Visual Simultaneous Localization and Mapping (SLAM) and Visual Odometry (VO) algorithms has led to the proposal of several accurate and efficient visual SLAM systems \cite{orb-slam3, air-slam}. A key focus of current research is applying visual SLAM to achieve long-term, stable localization for real-world mobile robots. Mobile robots are required to operate continuously in various seasons and under different weather conditions. Therefore, maintaining stable localization despite changes in lighting, seasonal variations, and other appearance changes has become a critical challenge. \par 

Map-based long-term visual localization system can be divided into three types: vector maps, object maps, and keypoint maps. Keypoint maps, which directly use results from SFM or vSLAM, offer the best versatility. However, handcrafted features have poor illumination robustness, and learning-based features are less computationally efficient. Vector maps are manually created using traffic signs and are commonly used for localization in urban roads and parking lots, especially in autonomous driving. Object maps build maps by identifying and estimating the positions and sizes of objects like tables, chairs, and trees for long-term localization. While both vector and object maps provide robust localization, their application is limited, and they are not universal solutions for all scenes.\par

\begin{figure}[!t]
        \centering
        \includegraphics[width=0.48\textwidth]{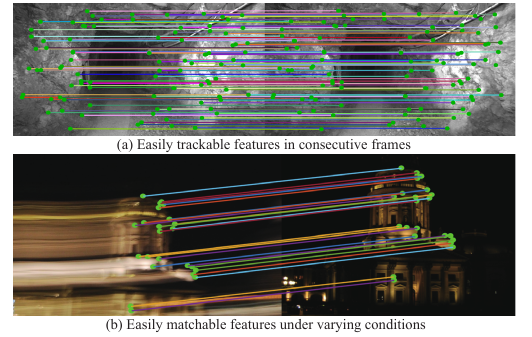}
        \caption{\textbf{An intuitive comparison of handcrafted and learned features.} (a) shows the matching results of ORB \cite{orb} features in a tunnel with repetitive textures, while (b) shows the matching results of D2-Net \cite{D2-net} under lighting variations. Features that are easy to track help maintain stable localization across consecutive frames, while features that are easy to match enable robust matching over long-term lighting changes.} 
        \label{fig_kp} 
        \end{figure}

Handcrafted and learned features have completely unified representations, which has led many works to attempt replacing handcrafted features with learned ones for improved long-term localization \cite{air-slam, hloc}. However, these methods struggle to balance efficiency and performance. This is because learned features focus more on repeatability and matching ability under varying lighting and viewpoints, while handcrafted features prioritize real-time efficiency and continuous tracking capability. This represents a fundamental difference: for example, local corner points in repeated textures can be tracked continuously but struggle with matching under large viewpoint changes, as show in Fig. \ref{fig_kp}. Therefore, we believe that learned features should not directly replace handcrafted features; instead, they should work together to achieve better long-term visual localization.

We build a keypoint map for long-term localization using two unified yet fundamentally different features, removing dependence on specific environments. We use handcrafted features for real-time relative pose estimation due to their efficiency and ability to track continuously under small viewpoint changes. Learned features, while less efficient, are used for low-frequency absolute pose computation because they excel at finding easily matchable locations. Through Handcrafted-Learned Feature Collaboration, we propose a hierarchical localization framework that enables pure visual localization across seasons and weather on a CPU. To support various learned feature types, including detect-then-describe and detect-and-describe paradigms, we propose a unified extraction framework that maintains compatibility with recent developments.

\begin{figure*}[!t] 
    \centering \includegraphics[width=\textwidth]{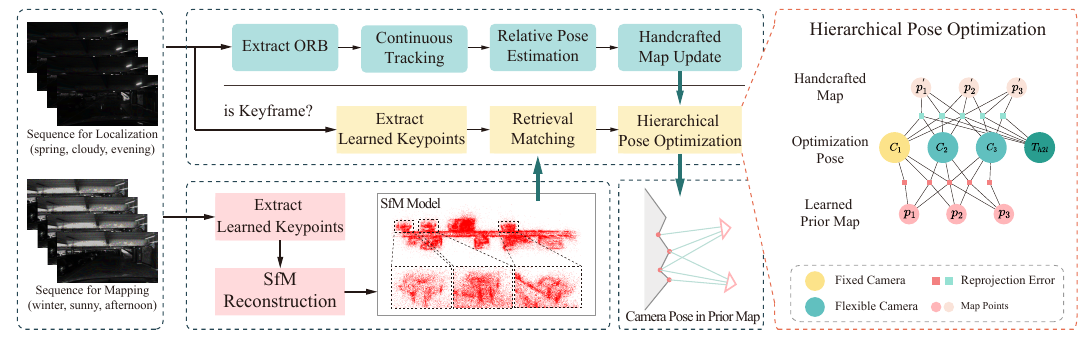}
    \caption{\textbf{Hierarchical Pipeline of Visual Localization.} Visual localization encompasses two primary phases: mapping and positioning. Initially, a conventional Structure-from-Motion (SfM) pipeline is employed to construct a learning-based feature map. Subsequently, multi-condition image sequences captured under varying seasonal and weather conditions are utilized for localization. Within the hierarchical localization framework, traditional ORB features facilitate continuous inter-frame tracking and relative pose estimation, enabling real-time construction of a handcrafted feature map. A subset of keyframes is then selected for learning-based feature extraction and subsequent matching with the prior map. The final positioning is achieved through an optimization process that minimizes reprojection errors between local keyframes and both handcrafted and learned feature maps, thereby determining the camera's precise location within the pre-established map.} 
    \label{pipeline} 
    \end{figure*}
    
In summary, the main contributions of this paper are as follows:

\begin{itemize}
  \item [1)] 
  We propose using two differentiated features with unified representations for long-term visual localization, balancing efficiency, performance, and environmental adaptability.
  \item [2)]
  We introduce a unified framework for extracting learned features, enhancing the system's long-term localization capabilities.
  \item [3)]
  We present a hierarchical pose optimization algorithm that simultaneously and efficiently refines handcrafted maps, learned prior maps, and multiple consecutive camera poses, achieving effective fusion of handcrafted-learned features.
\end{itemize}

\section{RELATED WORK}

Vision-based long-term localization methods can be classified into three categories: keypoint map-based localization, vector map-based localization, and object map-based localization. All these methods aim to address the challenges of handcrafted features' sensitivity to lighting and viewpoint changes.\par 

\subsection{Keypoint map-based localization}

Keypoint map methods offer environment-independent localization and can be divided into two main categories. The first category includes real-time localization algorithms that rely on handcrafted keypoints, often integrated with visual SLAM systems. For example, ORB-SLAM3 \cite{orb-slam3} uses handcrafted keypoint maps for localization. However, this approach is limited by the matching of handcrafted keypoints and becomes impractical when dealing with long-term appearance changes. The second category uses learned keypoints from offline 3D reconstruction for real-time localization \cite{hloc}. While this ensures long-term localization, it comes with high computational complexity and typically requires GPU hardware. Some works \cite{air-slam} have attempted to use learned keypoints in SLAM systems for long-term localization, but they also depend on powerful hardware. As of current knowledge, no universally applicable and efficient visual localization algorithm exists. \par 

\subsection{Vector map-based localization}

Vector maps are typically created using manually drawn ground traffic signs and are widely used for localization in urban roads and parking lots, especially in autonomous driving. These maps are most effective in environments with ground markings, such as city streets and parking garages. \cite{MonoLoc} first proposed using traffic signs on urban roads to create vector maps for localization. HDMI-Loc \cite{HDMI_Loc} introduced particle filters for localization within high-definition vector maps. AVP-Loc \cite{AVP_Loc} used a surround-view BEV perspective to segment and match ground markings, improving accuracy and robustness for long-term localization in parking garages.\par 

\subsection{Object map-based localization}

The object map method builds maps by repeatedly recognizing objects in the scene and estimating their positions and sizes. Common objects, such as tables, chairs, and trees, are used for long-term localization. \cite{ObjectLoc} first proposed modeling objects as ellipsoids to estimate camera poses. OA-SLAM \cite{OA-SLAM} developed a complete SLAM system that integrates camera localization, object map creation, and relocalization. ObVi-SLAM \cite{ObVi-SLAM} further applied recognition networks to achieve long-term, cross-seasonal localization and was successfully deployed on real-world robots. \par 

While vector maps and object maps have been successful in long-term visual localization, their applicability is limited. Vector maps are mainly used in parking lots and highways with clear road signage. Object maps are commonly applied in indoor environments with many known objects. In some industrial scenarios, additional training data specific to the target objects is needed to retrain recognition models. Therefore, neither approach provides a universally applicable solution for visual localization.\par

\section{METHOD}
\label{hyop}


This section first outlines the hierarchical localization framework, which integrates real-time tracking with asynchronous optimization. Next, we introduce a unified representation method to resolve feature heterogeneity between handcrafted and learned keypoints. Finally, the hierarchical pose optimization mechanism is comprehensively explained, achieving robust pose estimation through fusion of geometric and semantic information from dual-source maps.

\subsection{System Overview}
The proposed method is composed of two distinct and independent modules: offline mapping and online localization, as shown in Fig. \ref{pipeline}. 

\subsubsection{Mapping}

First, deep neural networks are used to extract discriminative and repeatable features from input images. These learned features, often more robust to variations in illumination, viewpoint, and texture, are then matched across multiple views to establish correspondences. With these correspondences, Structure-from-Motion (SfM) algorithms estimate camera poses and reconstruct the 3D structure of the scene through triangulation and bundle adjustment. By leveraging learning-based features, the reconstruction process can achieve greater consistency and completeness, especially in challenging scenarios where traditional handcrafted features may fail. \par 
The geometric reconstruction pipeline of SfM is well-established and user-friendly. In practice, COLMAP \cite{colmap} is employed to reconstruct a visual map of the environment. It is worth noting that any 3D reconstruction method or vSLAM algorithm can be utilized during the mapping stage.

\subsubsection{Real-Time Relative Pose Estimation}

To achieve real-time pose estimation, a tracking thread similar to that in ORB-SLAM3 \cite{orb-slam3} is used to determine the relative pose between image frames. First, handcrafted ORB \cite{orb} features are extracted using the OpenCV library. Then, these features are associated with those from previous image frames using the projection matching method described in Sec. \ref{sec_ma}. The matching results are used for estimating the relative pose between image frames, as outlined in Sec. \ref{sec_pe}, as well as for updating the depth of map points.

\subsubsection{Prior Map Alignment}
Only a small subset of carefully selected keyframes is used to establish associations with the prior map, ensuring efficient use of learned features. However, the discontinuous nature of feature matching significantly increases the difficulty of aligning images with the prior map. To address this, the relative pose estimated in real time is first used to predict the absolute position of the current keyframe. After extracting the learned features as Sec. \ref{sec_ex}, the projection matching method described in Sec. \ref{sec_ma} is applied again to associate the current keyframe with the prior map. After pose optimization as outlined in Sec. \ref{sec_pe}, projection matching is performed once more to obtain additional accurate associations. Finally, by applying the hierarchical pose optimization described in Sec. \ref{sec_hp}, both handcrafted and learned map observations are jointly optimized to estimate the current camera pose within the prior map.
\subsection{Unified Learning-based Feature Extraction}
\label{sec_ex}
\begin{figure}[!t] 
    \centering \includegraphics[width=0.49\textwidth]{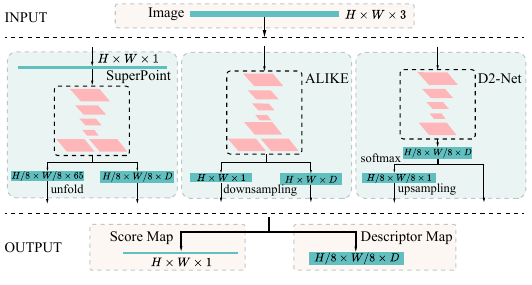}\caption{\textbf{Unified keypoint extraction process.} The networks for learning different keypoints are unified into standard input and output interfaces. The input is a color image of size $H\times W \times 3$, and the output consists of a score map of size $H\times W \times 1$, and a descriptor map of size $H/8 \times W/8 \times D$. The SuperPoint\cite{superpoint} network takes grayscale images as input, so a conversion is applied beforehand. Its output is a tensor of size $H/8\times W/8\times 65$, which needs to be processed via an unfold operation to achieve the standard form. The descriptor map of ALIKE\cite{alike} is downsampled to obtain the standard size. The score map of D2-Net\cite{D2-net} is obtained by applying a softmax operation on the descriptor map, followed by upsampling.} 
    \label{detection} 
    \end{figure}
Due to the differences in extraction and description methods for various learned keypoints, integrating them into a unified framework is challenging. To achieve a consistent representation for all learned keypoints, a generalized format is defined. As shown in Fig. \ref{detection}, although SuperPoint\cite{superpoint}, ALIKE\cite{alike}, and D2Net\cite{D2-net} utilize different network architectures, they can all be converted into this unified generalized representation. \par 
In this generalized representation, a color image of size $H \times W \times 3$ is input, producing a score map of size $H \times W \times 1$ and a descriptor map of size $\frac{H}{8} \times \frac{W}{8} \times D$. In the score map, higher scores indicate a greater likelihood of the corresponding location being a keypoint. The descriptor map encodes unique features for each position, which are used for image matching. \par 
The score map alone is insufficient for determining the exact keypoint locations in the image. To achieve a more evenly distributed set of keypoints, non-maximum suppression (NMS) is necessary. For efficient implementation, a method similar to the \textit{GoodFeaturesToTrack} function in OpenCV is used for keypoint extraction. First, the local maxima within a 3x3 neighborhood are retained. Then, we apply a maximum spacing sampling method to ensure an even distribution of the keypoints. 
\subsection{Projection Matching}
\label{sec_ma}
Due to the large number of map points in the map, matching each one individually with the current image is impractical. Therefore, it is essential to first project the map points into the image based on the estimated relative pose, and then perform feature association. This approach effectively reduces the complexity of matching between the image and the map, and is applied to both handcrafted and learned maps during image association. \par 
Specifically, since the proposed method establishes associations with the prior map using only a small number of keyframes, accumulated error may become significant. The sparsity of these associations makes it difficult to establish sufficient correspondences through a single round of matching. Therefore, the current frame and the prior map undergo a process of projection matching followed by pose optimization, which is then repeated. This iterative process helps to discover more reliable associations between the prior map and the keyframes. \par 

In the projection matching process, all prior map frames near the initially estimated position are first identified. Then, candidate map points are filtered based on the viewing angle of their associated observations. These map points are projected onto the pixel plane of the current frame, and potential matches are searched for in the surrounding regions. This approach effectively leverages the depth information embedded in the prior map and significantly improves the quality of feature associations. \par 

\subsection{Pose and Depth Estimation}
\label{sec_pe}

Bundle Adjustment is sensitive to initial values and prone to getting trapped in local minima. Therefore, directly combining handcrafted and learned observations into a single optimization process is not feasible. To obtain better initial estimates, the poses of image frames within the handcrafted map and the learned map are estimated separately. The handcrafted map is then updated in real time using depth estimation. These results are used as initial values for the joint optimization described in Sec. \ref{sec_hp}, ensuring the accuracy of the combined estimation.

Estimating the current frame’s pose by associating 3D map points with 2D image features is a classic Perspective-n-Point (PnP) problem. To improve the accuracy of pose estimation, we formulate it as a pose optimization problem rather than solving it through a direct linear method. This optimization-based approach is applied to estimate poses with respect to both the handcrafted and the learned maps. The formulation of the pose optimization problem is as
\begin{equation}
\begin{aligned}
    E &= \min_{\mathbf{\hat{T}}} \frac{1}{2} \sum_{i} \|\hat{e}_{i} \|^2 \\
      &= \min_{\mathbf{\hat{T}}} \frac{1}{2}\sum_{i} \|\mathbf{z}_{i} - \pi(\mathbf{\hat{T}} \mathbf{\hat{X}}_i)\|^2 ,
\end{aligned}
\end{equation}      
where $\hat{e}_i$ is the reprojection error corresponding to the $i$-th learned keypoint, $\mathbf{\hat{T}}$ is the pose of the current keyframe in the prior map, $\mathbf{z}_i$ is the pixel coordinates of the $i$-th learned keypoint, $\mathbf{\hat{X}}$ is the 3D coordinates of the map point in the prior map, and the function $\pi(\mathbf{X})$ projects the 3D vector in the camera coordinate system to the image coordinate system. The derivative of the error $\hat{e}_i$ of the $i$-th keypoint with respect to the camera pose $\mathbf{\hat{T}}$ in the prior map is given by
\begin{equation}
    \frac{\partial e_{i}}{\partial \delta \mathbf{\hat{T}}} = -\begin{bmatrix}
 \frac{f_x}{\hat{z}^{'}} & 0 & -\frac{f_x \hat{x}^{'}}{\hat{z}^{'2}} \\
 0 & \frac{f_y}{\hat{z}^{'}} & - \frac{f_y \hat{y}^{'}}{\hat{z}^{'2}}
\end{bmatrix} \begin{bmatrix}
 \mathbf{I} & -\mathbf{\hat{X}}_i ^{'\wedge } \\
\end{bmatrix},
\end{equation}
where $\delta \mathbf{\hat{T}}$ is the left perturbation of the pose $\mathbf{\hat{T}}$, and $\mathbf{\hat{X}}' = \mathbf{\hat{T}} \mathbf{\hat{X}}_i = [\hat{x}', \hat{y}', \hat{z}']^T$.

After identifying the keypoint locations, the subsequent step involves estimating the depth of the keypoints to calculate the positions of the map points. A coarse-to-fine strategy is employed to balance computational cost and accuracy. Initially, stereo images are used to compute the disparity between keypoints. By utilizing the disparity and the stereo baseline length $b$, the depth can then be derived as
\begin{equation}
    d = \frac{f_x b}{u_L-u_R},
\end{equation}
where $f_x$ represents the horizontal focal length of the camera, $u_L$ and $u_R$ are the positions of the keypoint in the left and right images, respectively, and $d$ is the estimated depth. However, due to the limitations of the baseline length, the accuracy of depth estimation remains relatively low. Therefore, refining the depth of keypoints using multi-view images and their pose estimates is essential for improving accuracy. \par 
Each keypoint is matched with the corresponding keypoints in adjacent frames to obtain as many matching results from different viewpoints as possible. Based on the keypoint locations obtained from multiple views, an optimization function that only optimizes the map point positions is constructed as 
\begin{equation}
\label{eq_depth}
\begin{aligned}
    E&=  \min_{\mathbf{X}} \frac{1}{2} \sum_{i,j} \|e_{ij}\| \\
    &= \min_{\mathbf{X}} \frac{1}{2}\sum_{i,j} \|\mathbf{z}_{ij} - \pi(\mathbf{T}_j \mathbf{X}_i)\|^{2} ,
\end{aligned}
\end{equation}
where $\mathbf{X}_i$ represents the 3D coordinates of the $i$-th map point, $\mathbf{T}_j$ denotes the extrinsic parameters of the $j$-th camera, and $\mathbf{z}_{ij}$ is the observed 2D pixel coordinate of the $i$-th map point as seen from the $j$-th camera. The projection function $\pi(\mathbf{X})$ transforms the 3D point $\mathbf{X}$ into its corresponding 2D image coordinate. To facilitate the optimization algorithm's solution, the computation of the error derivatives is essential. The derivative of the error $e_{ij}$ with respect to the map point position $\mathbf{X}_i$ is defined as

\begin{equation}
\label{div_x}
    \frac{\partial e_{ij}}{\partial \mathbf{X}_i}=-\begin{bmatrix}
 \frac{f_x}{z^{'}} & 0 & -\frac{f_x x^{'}}{z^{'2}} \\
 0 & \frac{f_y}{z^{'}} & - \frac{f_y y^{'}}{z^{'2}}
\end{bmatrix} \mathbf{R}_j,
\end{equation}
where $\mathbf{X}' = \mathbf{T}_j \mathbf{X}_i = [x', y', z']^T$, $f_x$ and $f_y$ are the camera's focal lengths, and $\mathbf{R}_j$ is the rotation matrix within the camera's extrinsic parameters $\mathbf{T}_j$.

\subsection{Hierarchical Pose Optimization}
\label{sec_hp}
\begin{figure}[!t] 
    \centering \includegraphics[width=0.49\textwidth]{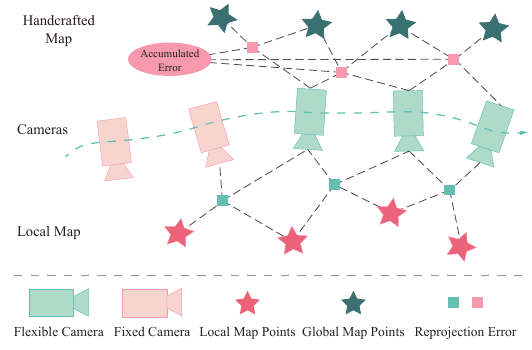}\caption{\textbf{Local BA optimization problem.} The dark green map points represent the prior visual map, which is also considered the global map. The pink map points represent the manually constructed real-time map. The light green cameras represent the camera poses near the current frame, referred to as the local map, whose poses will be optimized. The yellow cameras represent older cameras, which are fixed to provide continuity constraints. The projection errors of the global map points are used to estimate the accumulated error in the local map and the poses within the local map. The local map points are used solely to optimize the keyframe poses within the local map.} 
    \label{lba} 
    \end{figure}

The keyframe is associated with both the local map constructed from manual keypoints and the prior map constructed from learned keypoints. Therefore, the pose needs to be optimized using the reprojection errors from both maps. Since the localization task focuses only on the current pose, the positions of distant keyframes are less relevant. As a result, the most recent keyframes and their associated map points are constructed into a local map. Only the keyframes within this local map are optimized, and their poses are adjusted accordingly. As shown in Fig. \ref{lba}, all flexible frames and their associated map points are collectively referred to as the local map. Fixed frames, which are older frames not subject to optimization, serve the purpose of providing continuity constraints.\par 
Due to significant differences in viewpoint and appearance between the learned keypoint map and the keyframe, the number of correctly established associations can vary greatly. It is common to lose associations with prior map points in one or several consecutive keyframes. When the keyframe re-establishes associations with prior map points, substantial accumulated error may occur. At this point, fixed keyframes outside the local map retain the accumulated error, which cannot be optimized. Furthermore, these accumulated error propagate into the local map, making it difficult for the prior map to eliminate them effectively. \par 
To effectively use the prior map to eliminate accumulated error, we have designed a local BA algorithm, as shown in Fig. \ref{lba}. In this approach, manual map points are only used to solve for the position of the keyframe in the manual map, which contains accumulated error. On the other hand, we assume that all keyframes within the local map share the same accumulated error, denoted as $T_{\text{map}}$. The prior map points associated with all keyframes in the local map are primarily used to estimate this accumulated error. The advantage of this approach is that the accumulated error within the local map can be simultaneously estimated and corrected. \par 
In the local BA optimization algorithm, the reprojection errors of both manual keypoints and learned keypoints are simultaneously optimized. By minimizing their projection errors, the relative poses $\mathbf{T}_j$ between associated frames in the local map, the overall accumulated error $\mathbf{T}_{map}$ in the local map, and the positions of the map points $\mathbf{X}$ within the local map are obtained. Therefore, the loss function for this optimization problem is defined as
\begin{equation}
\begin{aligned}
    E &= \min_{\{\mathbf{T}, \mathbf{T}_{map}, \mathbf{X}\}} \frac{1}{2}  \sum_{j} \left( \sum_i \|e_{i,j} \|^2 + \sum_{l} \|\hat{e}_{l,j} \|^2 \right), \\
      &e_{i,j}= \mathbf{z}_{lj} - \pi(\mathbf{T}_j \mathbf{X}_i), \\
      &\hat{e}_{l,j} = \mathbf{z}_{ij} - \pi(\mathbf{T}_j^{-1} \mathbf{T}_{map}  \hat{\mathbf{X}}_l) ,
\end{aligned}
\end{equation}
where $e_{i,j}$ is the reprojection error of the manual keypoint, $\hat{e}_{i,j}$ is the reprojection error of the learned keypoint, $\mathbf{z}_{ij}$ is the pixel position of the $i$-th manual keypoint, $\mathbf{z}_{lj}$ is the pixel position of the $l$-th learned keypoint, $\mathbf{X}_i$ is the coordinates of the manual map point in the local map, $\mathbf{\hat{X}}_l$ is the coordinates of the learned map point in the prior map, and the function $\pi(\mathbf{X})$ projects the point $\mathbf{X}$ in the camera coordinate system onto the image. The derivatives of the error $e_{i,j}$ with respect to the camera extrinsics and map points, respectively. The derivative of the reprojection error $\hat{e}_{i,j}$ of the learned keypoints with respect to the accumulated error $\mathbf{T}_{map}$ in the local map is given by
\begin{equation}
    \frac{\partial \hat{e}_{lj}}{\partial \delta \mathbf{T}_{map}} = -\begin{bmatrix}
 \frac{f_x}{\hat{z}^{'}} & 0 & -\frac{f_x \hat{x}^{'}}{\hat{z}^{'2}} \\
 0 & \frac{f_y}{\hat{z}^{'}} & - \frac{f_y \hat{y}^{'2}}{\hat{z}^{'2}}
\end{bmatrix} \mathbf{R}_{j}^{-1} \begin{bmatrix}
 \mathbf{I} & -\hat{\mathbf{X}}^{'\wedge } \\
\end{bmatrix},
\end{equation}
where $\hat{\mathbf{X}}$ is a map point in prior map, $\hat{\mathbf{X}}^{'} = \mathbf{T}_j^{-1} \mathbf{T}_{\text{map}} \hat{\mathbf{X}}_l = [\hat{x}', \hat{y}', \hat{z}']^T$ is the map point in camera coordinate, $\mathbf{R}_j^{-1}$ is the rotation part of $\mathbf{T}_j$, and $\mathbf{X}^{\wedge}$ is the skew-symmetric matrix of the vector $\mathbf{X}$. The derivative of the reprojection error $\hat{e}_{i,j}$ of the learned keypoints with respect to the position of the keyframe in the local map is given by
\begin{equation}
    \frac{\partial \hat{e}_{lj}}{\partial \delta \mathbf{T}_j} = -\begin{bmatrix}
 \frac{f_x}{\hat{z}^{'}} & 0 & -\frac{f_x \hat{x}^{'}}{\hat{z}^{'2}} \\
 0 & \frac{f_y}{\hat{z}^{'}} & - \frac{f_y \hat{y}^{'2}}{\hat{z}^{'2}}
\end{bmatrix} \begin{bmatrix}
 \mathbf{I} & -\hat{\mathbf{X}}^{'\wedge } \\
\end{bmatrix},
\end{equation}
where $\hat{\mathbf{X}}$ refers to the map point in the prior map rather than the local map.

\section{Experiments}
In this section, we validate the effectiveness of the proposed hierarchical visual localization framework (VLOC) through a series of experiments. First, we demonstrate that learned keypoint extraction and matching methods cannot achieve real-time performance ($>$20Hz) on devices without a GPU, highlighting the necessity of the proposed hierarchical framework. Then, through comparative experiments with existing localization methods based on handcrafted keypoint maps and learned keypoint maps, we demonstrate the robustness, accuracy, and efficiency advantages of our approach in long-term, dynamically changing environments. Finally, by comparing localization performance across different keypoints, we verify the high adaptability of this framework to arbitrary learned keypoints.
\subsection{Learned keypoints efficiency}
In this experiment, we selected four typical edge devices representing different hardware platforms to test the efficiency of various learned keypoint extraction and matching methods. The NVIDIA Jetson AGX Orin 32GB, with 200 TOPS of AI computing power and a maximum GPU frequency of 1.2 GHz, represents NVIDIA GPUs. The Intel i5-1135G7, with a maximum turbo frequency of 4.2 GHz, serves as the CPU representative and was installed on a compact onboard computer to evaluate CPU inference efficiency. This processor also integrates Intel® Iris® Xe Graphics, allowing us to test iGPU performance up to 1.3 GHz with 80 execution units; both Experiments 2 and 3 used this setup. Lastly, the Rockchip RK3588, with 6 TOPS@INT8 NPU performance, was chosen as the NPU representative. \par 

To further optimize inference efficiency, we used platform-specific acceleration libraries. The TensorRT library was used on NVIDIA GPUs to maximize inference performance, while the Intel OpenVINO library was employed to accelerate inference on CPU and iGPU platforms. For the NPU, the Rockchip RKNPU library was applied to enhance inference efficiency. All inference times were calculated by averaging the results of 2000 inferences on 512×512 resolution images. \par 

Table \ref{inference} presents the average computation times for different types of learned keypoints on each platform. The results show a significant difference in performance between GPU and non-GPU devices. NVIDIA GPUs can meet the real-time requirement of over 20 Hz, while the Intel iGPU reaches a near-real-time level of 15 Hz. In contrast, the Intel CPU and NPU operate at approximately 10 Hz or even as low as 5 Hz, making it challenging to achieve higher real-time performance. We recorded the average operating frequency of different types of keypoints used for localization, including keypoint extraction, matching, and pose estimation, with all computations performed on a single CPU. This demonstrates that the proposed hybrid structure significantly enhances operational efficiency.\par 
\begin{table}[t]
        \begin{center}
                \caption{efficiency comparison}
                \label{inference}
                \setlength{\tabcolsep}{1mm}{
                \begin{tabular}{cccccc}
                \toprule
                \multirow{2}{*}{\textbf{Method}} & 
                \multicolumn{4}{c}{Interface time (ms) $\downarrow$} & 
                \multirow{2}{*}{\textbf{Frequency} $\uparrow$} \\
                \cmidrule{2-5} 
                ~ &
                \textbf{NVIDIA GPU}  &
                \textbf{iGPU} &
                \textbf{CPU} &
                \textbf{NPU}\\
                \midrule
                SuperPoint \cite{superpoint}  & 8.22 & \textbf{28.17} & 114.72 & 126.32 & 29 Hz\\
                ALIKE-T\cite{alike} & 49.95 & 70.04 & 94.81 & 755.34 & 22 Hz\\
                D2-Net \cite{D2-net} & 18.77 & 109.99 & 364.34 & 456.24 & / \\
                DISK \cite{disk} & 116.36 & 408.69 & 589.78 & 7374.04 & / \\
                XFeat \cite{xfeat} & \textbf{4.26} & 73.58 & \textbf{11.17} & \textbf{75.42} & 29Hz\\
            \bottomrule
        \end{tabular}}
    \end{center}
\end{table}

\subsection{Cross-seasonal visual localization}

\begin{table*}[t]
    \begin{center}
        \caption{Cross-seasonal visual localization comparison}
        \label{localization}
        \setlength{\tabcolsep}{0mm}{
        \begin{tabular}{ccccccccccc}
            \toprule
            \multirow{3}{*}{\textbf{Sequence}} &
            \multicolumn{5}{c}{ATE (m) $\downarrow$} & \multicolumn{5}{c}{RPE (m) $\downarrow$} \\
            \cmidrule(lr){2-6} \cmidrule(lr){7-11} 
             ~ & 
            \multirow{2}{*}{{ORB-SLAM3 \cite{orb-slam3}}} &
            \multirow{2}{*}{{AirSLAM\cite{air-slam}}} &
            \multicolumn{3}{c}{\textbf{Ours}} &
            \multirow{2}{*}{{ORB-SLAM3 \cite{orb-slam3}}} &
            \multirow{2}{*}{{AirSLAM\cite{air-slam}}} &
            \multicolumn{3}{c}{\textbf{Ours}} \\
            \cmidrule{4-6} \cmidrule{9-11}
            ~ & 
            ~ & ~ &{SuperPoint\cite{superpoint}} & 
            {ALIKE\cite{alike}} &
            {XFeat\cite{xfeat}} &
            ~ & ~ &
            {SuperPoint\cite{superpoint}} & 
            {ALIKE\cite{alike}} &
            {XFeat\cite{xfeat}} \\
            \midrule
            {parking\_garage\_1} & {11.13} & {23.71} & \textcolor{red}{4.79} & \textcolor{green}{5.35} & {8.59} & \textcolor{red}{0.03} & {8.00} & {0.05} & \textcolor{green}{0.05} & {0.06} \\
            {parking\_garage\_2} & {7.18} & {55.84} & {4.21} & \textcolor{red}{3.73} & \textcolor{green}{4.04} & \textcolor{red}{0.04} & {75.54} & {0.12} & {0.10} & \textcolor{green}{0.09} \\
            {parking\_garage\_3} & {5.82} & {26.15} & \textcolor{green}{4.81} & {5.94} & \textcolor{red}{4.64} & \textcolor{green}{0.04} & {4.08} & {0.06} & \textcolor{red}{0.27} & {0.06} \\
            \bottomrule
        \end{tabular}}
    \end{center}
\end{table*}

\begin{figure}[!t] 
    \centering \includegraphics[width=0.49\textwidth]{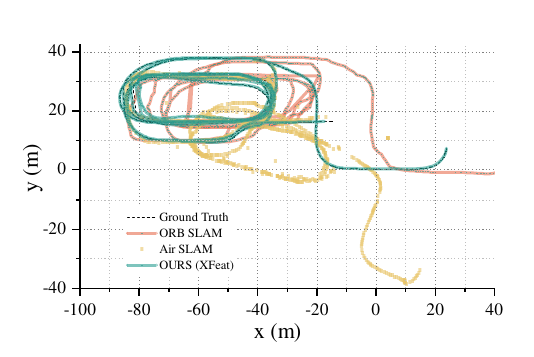}\caption{\textbf{Comparison of localization trajectories across seasons.} The comparison of localization trajectories obtained from handcrafted keypoint maps, learned keypoint maps, and the proposed hierarchical localization method intuitively demonstrates the effectiveness of the proposed approach. Notably, the localization results provided by AirSLAM \cite{air-slam} are unavailable at certain moments, resulting in discrete and non-continuous trajectories.} 
    \label{traj} 
    \end{figure}

We compared our method with existing localization methods on three "Parking Garage" sequences from the 4Seasons \cite{4Seasons} dataset. In these sequences, the vehicle repeatedly circulates within a three-level parking structure, where significant lighting variations between sequences present greater challenges for localization algorithms. The 4Seasons \cite{4Seasons} dataset encompasses seasonal variations and challenging perception conditions encountered in autonomous driving, covering environments such as urban areas, multi-level parking garages, rural settings, and highways. Additionally, it provides globally consistent reference poses obtained through the fusion of direct stereo visual-inertial odometry and RTK-GNSS, making it ideal for testing the localization performance of algorithms in complex and dynamic scenarios. \par


In this experiment, two representative localization algorithms were used for comparison. ORB-SLAM3 \cite{orb-slam3} is a mature and widely-used visual SLAM system based on handcrafted keypoints. It achieves robust real-time localization in both indoor and outdoor environments through efficient loop closure and multi-map support. AirSLAM\cite{air-slam}, on the other hand, is a novel visual SLAM algorithm that combines deep learning with traditional backend optimization to tackle the challenges of lighting variations. With its lightweight design and acceleration framework, AirSLAM \cite{air-slam} runs efficiently on embedded platforms. Both algorithms support a pure localization mode, enabling fast and efficient localization on pre-existing maps without requiring remapping. \par 

The evo\cite{evo} tool was used to evaluate trajectory accuracy. After aligning the estimated trajectory with the ground truth using SE(3) Umeyama alignment,  Absolute Trajectory Error (ATE) and Relative Pose Error (RPE) metrics \cite{ape_evo} were calculated to assess trajectory accuracy. The ATE metric reflects the system's overall localization accuracy, while the RPE metric assesses accuracy over the local trajectory. Table \ref{localization} presents the comparative results of the three algorithms, with the optimal result highlighted in red and the second-best in green. The ATE results clearly show that the proposed method significantly improves global localization accuracy in dynamic environments, while the RPE results indicate a slight reduction in local accuracy. Taking Sequence 1 from the dataset as an example, the trajectory results of the three algorithms are shown in Fig. \ref{traj}. ORB-SLAM3 \cite{orb-slam3} suffers from severe drift due to cumulative error during long-term pure localization in complex environments. The Air-SLAM \cite{air-slam} method shows weaker stability, achieving high localization accuracy in the first half of the trajectory but exhibiting noticeable drift in the second half. In contrast, our proposed method maintains efficient performance while delivering globally consistent localization, demonstrating substantial robustness and accuracy advantages in challenging environments.

\subsection{Visual localization with different learned keypoints}

\begin{table*}[t]
        \begin{center}
                \caption{Localization performance of different keypoints}
                \label{keypoints}
                \setlength{\tabcolsep}{0.5mm}{
                \begin{tabular}{cccccccccc}
                \toprule
                \multirow{2}{*}{\textbf{Sequence}} & \multirow{2}{*}{\textbf{Label}} &
                \multicolumn{4}{c}{ATE (m) $\downarrow$} & \multicolumn{4}{c}{RPE (m) $\downarrow$} \\
                \cmidrule(lr){3-6} \cmidrule(lr){7-10} 
                ~ & ~ & 
                \textbf{ORB\cite{orb}}  &
                \textbf{SuperPoint \cite{superpoint}} &
                \textbf{ALIKE \cite{alike}} &
                \textbf{XFeat \cite{xfeat}} &
                \textbf{ORB\cite{orb}}  &
                \textbf{SuperPoint\cite{orb}} &
                \textbf{ALIKE\cite{alike}} &
                \textbf{XFeat\cite{xfeat}}\\
                \midrule
                {office\_loop\_1} & {spring$,$ sunny$,$ afternoon} & 96.92 & \textcolor{green}{2.83} & \textcolor{red}{2.33} & 8.86 & 0.11 & \textcolor{red}{0.06} & \textcolor{green}{0.10} & 0.21 \\
                {office\_loop\_2} & {spring$,$ sunny$,$ afternoon} & 32.85 & 21.98 & \textcolor{green}{6.33} & \textcolor{red}{3.02} & 0.17 & \textcolor{green}{0.17} & 0.24 & \textcolor{red}{0.09} \\
                {neighborhood\_1} & {spring$,$ cloudy$,$ afternoon} & 22.74 & 0.38 & \textcolor{red}{0.35} & \textcolor{green}{0.35} & 0.03 & 0.02 & \textcolor{green}{0.02} & \textcolor{red}{0.02} \\
                {neighborhood\_2} & {fall$,$ cloudy$,$ afternoon} & 5.38 & \textcolor{green}{3.79} & \textcolor{red}{2.83} & 3.84 & \textcolor{red}{0.04} & 0.08 & \textcolor{green}{0.04} & 0.15 \\
                {neighborhood\_3} & {fall$,$ rainy$,$ afternoon} & {5.37} & \textcolor{red}{3.51} & \textcolor{green}{3.59} & {4.27} & \textcolor{red}{0.05} & \textcolor{green}{0.06} & {0.12} & {0.17} \\
                {neighborhood\_4} & {winter$,$ cloudy$,$ morning} & {13.48} & \textcolor{red}{5.62} & \textcolor{green}{5.62} & {5.84} & \textcolor{green}{0.06} & {0.07} & \textcolor{red}{0.04} & {0.12} \\
                {neighborhood\_5} & {winter$,$ sunny$,$ afternoon} & \textcolor{red}{2.34} & {3.65} & \textcolor{green}{2.40} & {2.49} & \textcolor{red}{0.02} & {0.04} & \textcolor{green}{0.03} & {0.03} \\
                {neighborhood\_6} & {spring$,$ cloudy$,$ evening} & {10.45} & {5.35} & \textcolor{green}{3.24} & \textcolor{red}{3.15} & \textcolor{red}{0.03} & {0.11} & {0.06} & \textcolor{green}{0.04} \\
                {neighborhood\_7} & {spring$,$ cloudy$,$ evening} & {5.49} & {3.79} & \textcolor{green}{3.09} & \textcolor{red}{3.04} & \textcolor{red}{0.02} & {0.05} & \textcolor{green}{0.04} & {0.04} \\     
            \bottomrule
        \end{tabular}}
    \end{center}
\end{table*}

\begin{figure*}[!t] 
    \centering \includegraphics[width=\textwidth]{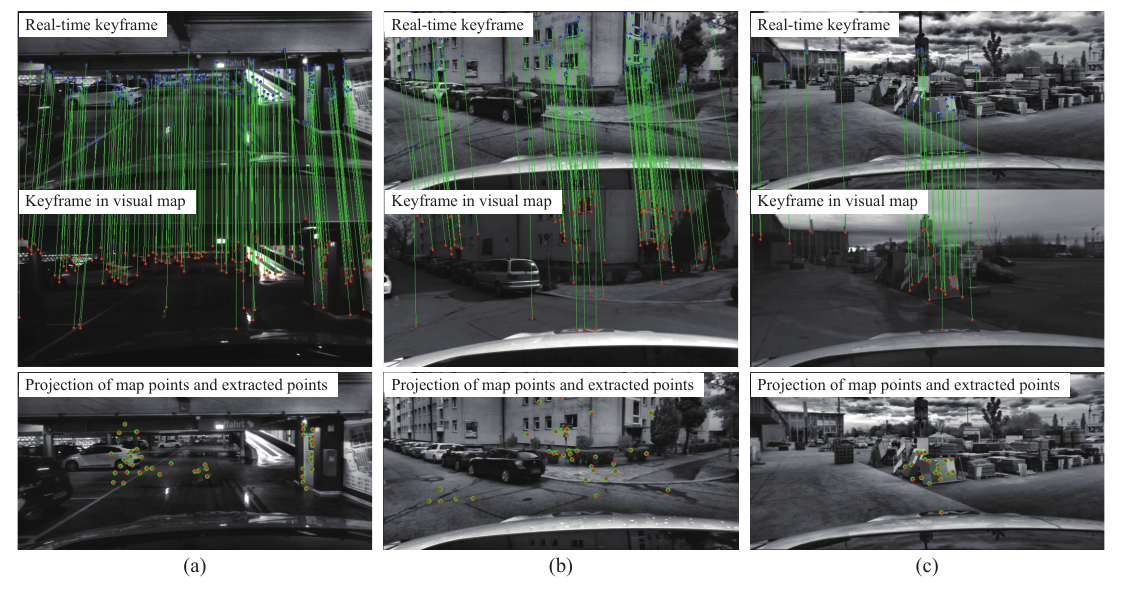}
    \caption{\textbf{Visualization of Learned Keypoint Maps for Localization.} The first row of images shows the keyframes selected by the real-time localization algorithm, with the extracted learned keypoints marked in blue. The second row displays images of keypoint maps constructed under different seasonal and weather conditions, where the learned keypoints are marked in red. The matching relationships between real-time keyframes and keypoints in the prior map are highlighted in green lines, demonstrating the strong matching capability of the learned keypoints. In the third row, the projection locations of visual map points and the real-time extracted keypoints are shown in green and yellow, respectively. This indicates that the estimated pose of the current frame remains consistent with the keypoint map.} 
    \label{match} 
    \end{figure*}

In this experiment, the Office Loop and Neighborhood sequences from the 4Seasons \cite{4Seasons} dataset were used to conduct localization experiments across various seasons and time periods. The specific weather conditions and time periods for each sequence are detailed in Table \ref{keypoints}. \par 

Each set of image sequences underwent three repeated experiments under the same configuration, with the results averaged. As shown in Table \ref{keypoints}, the optimal and second-best values for each experiment are indicated in red and green, respectively. For the APE, the hierarchical framework significantly improved global accuracy across most datasets by using learned keypoints to refine localization results, particularly in complex scenes. The only exception was the Neighborhood\_5 dataset, where the lower scene difficulty allowed for high localization accuracy using only handcrafted keypoints, resulting in no significant additional improvement from the hierarchical framework. Regarding RPE, the lower frequency of localization corrections meant that non-corrected phases relied on handcrafted keypoints, which could temporarily decrease local accuracy during correction moments, leading to an increase in RPE. Fig. \ref{match} presents a visualization of intermediate results from hierarchical map localization. The first two rows demonstrate the cross-season matching capability of learned keypoints. The third row, by visualizing reprojection distances, verifies the accuracy of pose estimation based on the prior visual map. \par 

Through a systematic analysis of the experimental results, we validated the localization performance advantages of the proposed hierarchical framework across different keypoint types. The framework significantly improved global localization accuracy, particularly in complex scenes with substantial appearance variations. Although the local error (RPE) increased due to a lower correction frequency, the overall results indicate that this framework effectively balances real-time performance and accuracy. It demonstrates enhanced robustness and stability in complex, dynamic environments.

\section{Conclusion}
Considering the complementary and contradictory characteristics of handcrafted and learned features, we propose a general and efficient long-term visual localization framework that supports integration with any state-of-the-art learned features. Unlike previous approaches that aim to replace handcrafted features entirely with learned ones, we believe that learned features are more suitable for matching, while handcrafted features excel in tracking. Therefore, both should be jointly utilized to achieve robust long-term localization. By designing a holistic system and a hierarchical map optimization algorithm, we have achieved real-time long-term localization on a CPU, effectively validating the proposed hybrid feature integration strategy. Looking ahead, we anticipate that our method will be further applied in real-world industrial robotic systems to enable long-term stable visual localization and navigation. Additionally, we plan to explore the possibility of achieving better localization performance through a unified feature representation, learn trackability within images, and investigate more efficient network architectures and training methods for learned feature extraction.
\bibliographystyle{IEEEtran}
\bibliography{IEEEabrv,bibtex/IEEEabrv}
\vfill

\end{document}